\documentclass[10pt,twocolumn,letterpaper]{article}

\usepackage{cvpr}              
\usepackage[accsupp]{axessibility} 

\usepackage[dvipsnames]{xcolor}

\usepackage{graphicx}
\usepackage{amsmath}
\usepackage{amssymb}
\usepackage{times}
\usepackage{epsfig}
\usepackage{bm}
\usepackage{amsthm}
\usepackage{subcaption}
\usepackage{url}
\usepackage{gensymb}
\usepackage{ulem}
\usepackage[symbol]{footmisc}
\usepackage{makecell}
\usepackage{xcolor}
\usepackage{float}
\usepackage{tabularx}
\usepackage{multirow}

\definecolor{DeltaColor}{rgb}{0.039,0.73,0.71}
\definecolor{SetaColor}{rgb}{0.867, 0.0235, 0.376}
\definecolor{SigmaColor}{rgb}{0.98,0.45,0.0}
\definecolor{RedColor}{rgb}{0.8,0,0}
\definecolor{AlphaColor}{rgb}{0,0,0.8}
\definecolor{BetaColor}{rgb}{0.8,0,0.8}
\definecolor{GammaColor}{rgb}{0.5,0,0.7}
\definecolor{EpsilonColor}{rgb}{0.353,0.725,0.906}
\definecolor{TauColor}{rgb}{0.423,0.235,0.192}

\newcommand{\image}{X}
\newcommand{\mask}{M}

\newcommand{\Real}{\mathbb{R}}
\newcommand{\Integer}{\mathbb{Z}}
\newcommand{\codebook}{\mathbf{C}}
\newcommand{\token}{\mathbf{z}}
\newcommand{\tlabel}{\mathbf{y}}
\newcommand{\Labels}{\mathbf{Y}}

\definecolor{cvprblue}{rgb}{0.21,0.49,0.74}
\usepackage[pagebackref,breaklinks,colorlinks,citecolor=cvprblue]{hyperref}

\def\paperID{6100} 
\def\confName{CVPR}
\def\confYear{2024}

\title{Don't Look into the Dark: Latent Codes for Pluralistic Image Inpainting}

\author{Haiwei Chen, Yajie Zhao\\
	University of Southern California\\
    USC Institute for Creative Technologies\\
{\tt\small \{chenh,zhao\}@ict.usc.edu}
}

\begin{document}
\twocolumn[{%
\renewcommand\twocolumn[1][]{#1}%
\maketitle
\begin{center}
    \centering
    \captionsetup{type=figure}
    \includegraphics[width=0.95\textwidth]{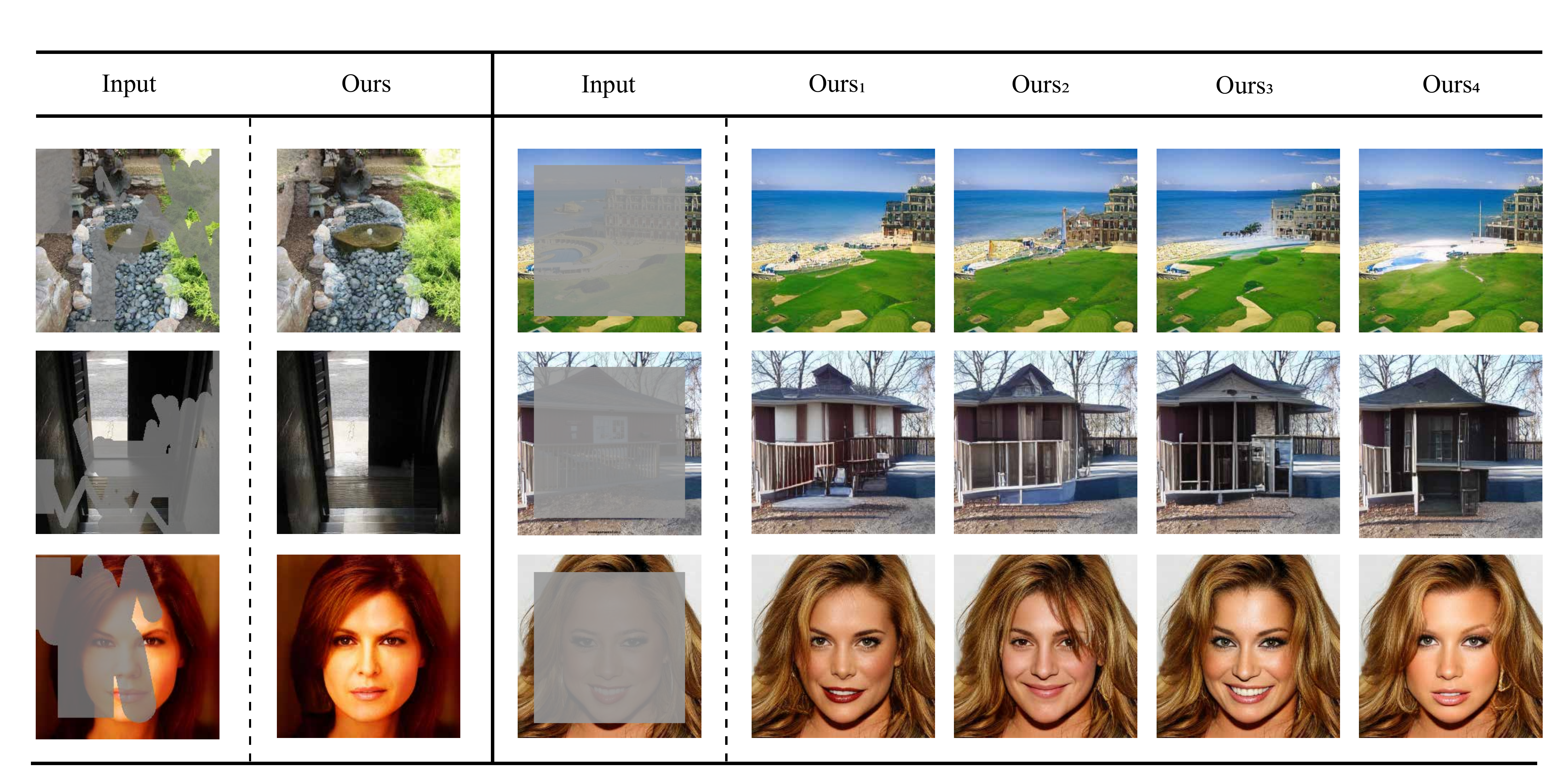}
    \captionof{figure}{Inpainting results on the Places Dataset~\cite{zhou2017places} (first two rows) and the CelebA-HQ Dataset~\cite{karras2018progressive} (third row). Our method is able to diversely complete partial image with free-form, large holes with state-of-the-art visual quality.}
    \label{fig:teaser}
\end{center}}]


\begin{abstract}
We present a method for large-mask pluralistic image inpainting based on the generative framework of discrete latent codes. Our method learns latent priors, discretized as tokens, by only performing computations at the visible locations of the image. This is realized by a restrictive partial encoder that predicts the token label for each visible block, a bidirectional transformer that infers the missing labels by only looking at these tokens, and a dedicated synthesis network that couples the tokens with the partial image priors to generate coherent and pluralistic complete image even under extreme mask settings. Experiments on public benchmarks validate our design choices as the proposed method outperforms strong baselines in both visual quality and diversity metrics. 
\end{abstract}   
\section{Introduction}
\label{sec:intro}

Image inpainting is the task of filling the missing pixels of a masked image with appropriate contents that are coherent to its visible regions. As a long-studied topic in computer vision, image inpainting has evolved from a restoration technique solely relying on existing information from the input image (e.g. \cite{bertalmio2000image}) to data-driven generative methods (e.g. \cite{zheng2019pluralistic, yu2019free, yi2020contextual, li2022mat, suvorov2021resolution, lugmayr2022repaint}) that hallucinates detailed contents from not only the observable pixels but also learned, rich image priors. 






  
 Pluralistic inpainting refers to the ability of a model to generate multiple plausible results that complete a partial image. It offers a view of image inpainting as a generative method that models the smooth distributions of the complete images given the partial image as prior information~\cite{zheng2019pluralistic}. However, modeling such distributions is challenging in the typical encoder-decoder network structures. In order to synthesize missing contents that both respect the partial image and maintain sample diversity, the decoder in this setting takes as input two types of information: 1) features propagated from the visible regions and 2) random noise vectors sampled from a prior distribution. If the training objective is to reconstruct a ground-truth image from a partial image, the objective itself may discourage conditioning on the random variable. Moreover, as the training dataset contains numerous examples that only require low-level information to complete an image (e.g. smoothly interpolating a wall texture), the model may choose to ignore the latent priors when the available image cues are strong enough to provide an answer. The phenomenon has been found in image translation networks~\cite{isola2017image}, where adding noise to generate a conditional image does little to create pluralistic results. 

 In this paper, we investigate a unique approach to pluralistic image inpainting, following a branch of recently developed synthesis methods known as the generative transformers~\cite{esser2021taming, razavi2019generating, chang2022maskgit}. These approaches synthesize images by procedurally predicting latent codes, termed as ``tokens'', that semantically encode information of an image, analogous to sentence generation in natural language processing~\cite{vaswani2017attention, devlin2018bert}. As tokens drawn in each step are based on a learned posterior distribution of the previous step, generative transformers offer fine-grained control over diversifying the synthesized contents. Our paper is the first to adapt the generative transformer framework to target the image inpainting task. To this end, we design a three-stage pipeline to 1) encode a partial image into discrete latent codes, 2) predict the missing tokens with a bidirectional transformer, and 3) couple the predicted tokens with the partial image priors and decode them into a complete image. The resulted method design fundamentally differentiates itself from typical inpainting approaches in the past: instead of modeling the complex interaction between the missing and observed regions, our method barely looks at the missing regions during the encoding and token prediction stage: the designed mask-aware encoder utilizes restrictive convolutions to only operate on the visible and near-visible regions, and the bidirectional transformer only attends to the visible tokens to make a prediction. This separation design leads to a paradigm that divides feature reasoning and generative modeling into two separate stages, which we found beneficial for large-mask pluralistic image inpainting.

 Through experiments, we have validated that our design choices lead to a robust, high-quality, and pluralistic solution to challenging image inpainting settings. Our method has achieved state-of-the-art performance both in terms of visual quality and sample diversity in two public benchmarks Places~\cite{zhou2017places} and CelebA-HQ~\cite{karras2018progressive}. 



\begin{figure*}[t]
  \centering
  \includegraphics[width=1.0\linewidth]{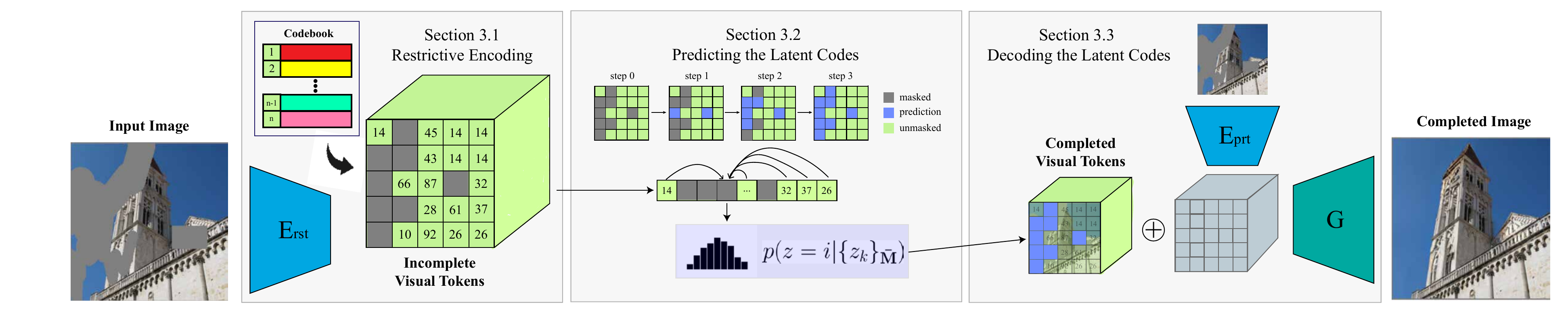}
  \caption{Overall pipeline of our method. $E_{rst}$ denotes our proposed restrictive encoder that predicts partial tokens from the source image (see Section~\ref{sec:encoder}). The grey square space in the figure denotes missing tokens, which are iteratively predicted by a bidirectional transformer (see Section~\ref{sec:transformer}). $E_{prt}$ denotes an encoder with partial convolution layers, which processes the source image into complementary features to the predicted tokens. The coupled features are decoded into a complete image by a generate $G$ (see Section~\ref{sec:decoder}).}
  \label{fig:network}
\end{figure*}

\section{Related Works}
\label{sec:related}

The term ``image inpainting'' was originally introduced in~\cite{bertalmio2000image} as a propagation algorithm that fills in the missing regions following lines near the boundaries. Following works are methodologically divided into two branches: the diffusion-based methods~\cite{fadili2009inpainting, ballester2001filling, shen2002mathematical, levin2003learning, sun2005image} that aim at interpolating information in the hole areas under local or global optimization constraints, and the patch-based methods~\cite{lee2016laplacian, darabi2012image, barnes2009patchmatch, criminisi2003object, drori2003fragment} that composite candidate patches from selected regions into the hole areas. The early works are characterized by analyzing the inner statistics of the single image provided, many of which have shown satisfactory results for filling small holes, restoring relatively simple salient structures, and reconstructing textures. 

Image inpainting has been advanced greatly by recent works that utilize artificial neural networks (ANN), as they introduce deep image priors to the synthesis of missing contents in a more robust and versatile way. Notably, the perceptual loss~\cite{johnson2016perceptual, zhu2017toward} and the adversarial loss~\cite{goodfellow2014generative} have been found to drastically improve the visual quality achieved by the ANN methods. Many works have proposed designs that incorporate both local coherence and global consistency through multi-scale discriminators~\cite{iizuka2017globally}, feature rearrangement~\cite{yan2018shift, song2018contextual}, spectral convolution~\cite{suvorov2021resolution}, attention layers~\cite{yu2018generative, liu2019coherent, zheng2022bridging, li2022mat, shamsolmoali2023transinpaint} and GAN inversion~\cite{yildirim2023diverse}. Mask-aware operations have also been explored in various forms, notably with partial convolution~\cite{liu2018image}, gated convolution~\cite{yu2019free} and continuously masked transformer~\cite{ko2023continuously}. Another relevant direction that shares similarity to our generative method is the progressive approach~\cite{guo2019progressive, li2020recurrent, zhang2018semantic, zeng2020high}, where inpainting is done step-wise via various forms of residual networks to gradually update both the mask and the feature maps of interest.  


While the majority of inpainting works (e.g. \cite{suvorov2021resolution, zheng2022bridging, pathak2016context, yu2018generative, song2018contextual, liu2019coherent, yan2018shift, iizuka2017globally, xiong2019foreground, liu2018image, yi2020contextual, yu2019free}) produce deterministic results given a fixed partial image input, various techniques have been proposed for pluralistic inpainting. \cite{zheng2019pluralistic} first studies the coupling of smooth latent priors with pixels generated for a masked image, where the authors suggest the undesirable side effect of reconstructive inpainting on sample diversity and propose a dual-path pipeline to better condition the generated results on both the observed regions and the sampled priors. \cite{liu2021pd} modulates a latent vector in the image space based on a coarse inpainting prediction. More recent works \cite{li2022mat, lugmayr2022repaint} have shown major improvements in the visual quality of pluralistic inpainting. MAT~\cite{li2022mat} achieves state-of-the-art inpainting quality with a unifying design of normalized transformer, shifting attention and style code modulation. RePaint~\cite{lugmayr2022repaint} leverages the denoising diffusion probabilistic model to image inpainting. The diffusion model can create realistic and diverse inpainting results through iteratively denoising resampled pixels in many steps. The limitation of it is the very slow sampling speed, which may be hindering for some real-world applications. 

Our method extends from a family of methods characterized by learning priors from discrete latent codes that are obtained from a vector-quantized autoencoder~\cite{van2017neural}. Past research in this direction has only focused on image synthesis, from directly predicting pixels as word tokens~\cite{chen2020generative}, to predicting tokens encoding visual features of larger receptive fields~\cite{van2017neural, esser2021taming}. While the pioneering works infer latent codes autoregressively, MaskGIT~\cite{chang2022maskgit} finds it beneficial to synthesize an image in a scattered manner with a bidirectional transformer: in every iteration, a number of new codes are predicted in parallel and inserted into scattered locations of the code map until the entire grid is filled. While~\cite{chang2022maskgit} has partially adapted its bidirectional framework to the image inpainting setting, our method design addresses several unanswered aspects of this adaptation: how partial images can be robustly masked into latent codes, and how the latent codes should be decoded into synthesized pixels that respect the observable area.






\section{Method}
\label{sec:method}

Our method is divided into three stages to complete an input partial image. The neural network model takes as input a partial image $\image_M$ and a mask image $\mask$ specifying the area to complete. The first stage encodes the partial image into a set of discrete tokens, referred to as latent codes, at a lower resolution and specifies the masked tokens that need to be predicted (Section~\ref{sec:encoder}); the second stage utilizes a bidirectional transformer to predict the missing tokens iteratively (Section~\ref{sec:transformer}); and the third stage couples the predicted tokens with features from the partial image and decodes them into a completed image (Section~\ref{sec:decoder}). Figure~\ref{fig:network} provides a visualization of the overall pipeline of our method.

\subsection{Encoding with Restrictive Convolutions}
\label{sec:encoder}

Our latent codes are represented by a discrete codebook $\codebook$ of learned tokens $\codebook = \{\token_k\}_{k=1}^{K} \in \Real^{n_z}$, where $k$ is the number of tokens and $n_z$ is the number of channels of each token feature. The tokens are given a set of labels $\Labels = \{\tlabel\}_{k=1}^{K} \in \Integer$. We employ the setup of VQGAN~\cite{esser2021taming} to learn the codebook by training on full images, where the token grid is $1/16$ of the resolution of the image. Ideally, encoding a partial image amounts to extracting valid tokens for the observed parts and invalid tokens, which are given a special \small [MASK] \normalsize token, for the masked parts. However, the convolutional nature of the VQGAN network leads individual tokens to encode not just local information, but also information of its proximity. Directly encoding partial images with a VQGAN encoder thus leads to degradation of the tokens, as the masked region inevitably affects how an encoder chooses to extract tokens. Another concern is the fact that pixel-level masking (e.g. on a 256x256 image) does not directly translate to token-level masking (on a 16x16 token grid): small regions of masked pixels may still contain rich information in their neighborhood. However, directly down-sampling the mask image may lead many observable pixels to be masked out at a lower resolution, therefore discarding a fair amount of useful information. A good approach is thus needed to determine when a token should be considered masked in the encoding step. 

\textbf{Separating Masked Pixels} The principle idea behind our encoding method is to prevent the participation of large areas of masked pixels in each convolutional network layer, controlled by a hyperparameter ratio $\alpha$. To simplify the explanations, let's consider an encoder network with only non-strided convolutions and down-sampling layers. The \textbf{standard partial convolution}~\cite{liu2018image} widely used in past image inpainting works are characterized by scaling the matrix multiplication in the convolution operation and an update rule for the mask:

\begin{equation}  
    \label{eq:partial}
    x' = \begin{cases} W^T(\mathbf{X} \bigodot \mathbf{\mask} ) \frac{1}{\text{sum}(\mathbf{\mask})} + b, & \text{if sum}(\mathbf{\mask}) > 0 \\
                       0, & \text{otherwise.}
    \end{cases} 
\end{equation}
\begin{equation}
    \mathbf{\mask}(x)' = \begin{cases} 1, & \text{if sum}(\mathbf{\mask}) > 0 \\
                       0, & \text{otherwise,}
    \end{cases}
\end{equation}

where $\mathbf{X}$ denotes a $n \times n$ neighborhood in the feature map under a convolution kernel size of $n$, and $\mathbf{\mask}$ denotes the $n \times n$ mask in that area. $\bigodot$ denotes element-wise multiplication. Effectively, the standard partial convolution mitigates the impact of masked pixels on the features' signal strength with an adaptive scaling and propagates new features into the masked pixels as long as there are visible pixels surrounding them. In contrast, we choose to separate the image prior learning step and the synthesis step aggressively in two different stages. In the encoding stage, we thus propose the \textbf{restrictive partial convolution} that only considers regions surrounded by a certain proportion of visible pixels:


\begin{equation}
    \label{eq:restrictive}
    x' = \begin{cases} W^T(\mathbf{X} \bigodot \mathbf{\mask} ) \frac{1}{\text{sum}(\mask)} + b, & \text{if} \: \frac{\text{sum}(\mathbf{\mask})}{\text{sum}(\mathbf{1})} >=\alpha \\
                       0, & \text{otherwise,}
    \end{cases} \\
    \end{equation} 
    
and at each down-sampling layer, we update the mask by:
\begin{equation}
    \label{eq:restrictive2}
    \mathbf{\mask}(x)' = \begin{cases} 1, & \text{if} \: \frac{\text{sum}(\mathbf{\mask})}{\text{sum}(\mathbf{1})} >=\alpha \\
                       0, & \text{otherwise,}
    \end{cases} \\
\end{equation}

where $\mathbf{1}$ denotes a $n \times n$ constant tensor of value 1. Different from the standard partial convolution, the algorithm does not update the mask at each restrictive partial convolution layer. The changes made here prevent feature propagation to densely unmasked regions subject to the $\alpha$ value. 

\begin{figure}[t]
  \centering
   \includegraphics[width=0.8\linewidth]{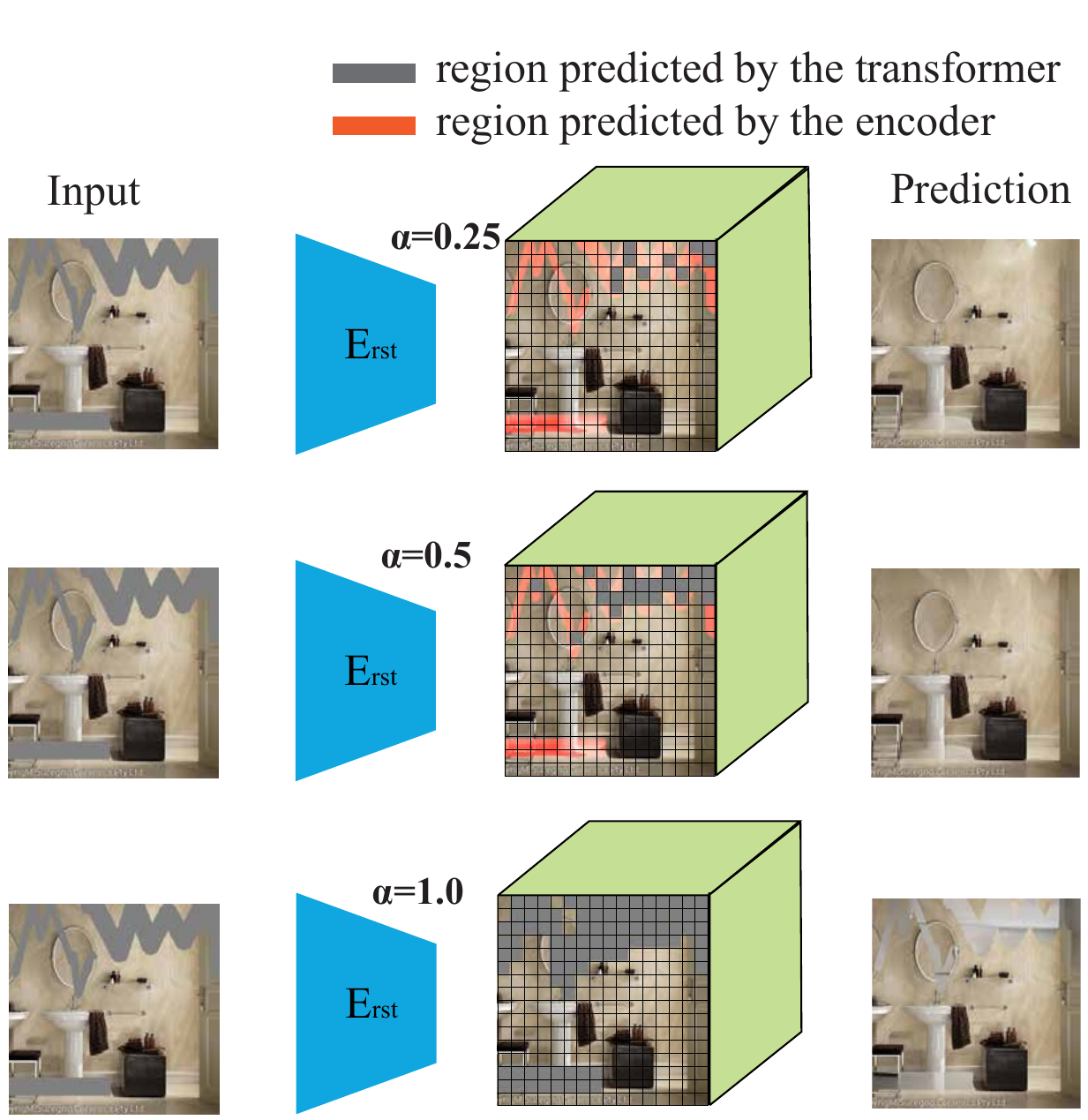}
   \caption{A visualization of mask down-sampling, shown on a 16x16 grid on the third column, from different $\alpha$ values following Equation~\ref{eq:restrictive2}. Smaller $\alpha$ values (top two rows) lead the restrictive encoder to predict tokens for more small mask areas (marked by the red pixels). Larger $\alpha$ is undesirable (bottom two rows) as it unnecessarily discards useful information from the image, leading to more inconsistent inpainting results.}
   \label{fig:alpha}
\end{figure}

Besides separating the masked regions from the observed features, this particular convolution design also addresses the inevitable mismatch between the input pixel-level mask and the updated mask seen in the much-lower-resolution feature space. By using a small $\alpha$ value, the encoder is designated to fill in tokens for small regions of unseen pixels while leaving the larger regions to be predicted in the next stage (see Section~\ref{sec:transformer}).  Figure~\ref{fig:alpha} provides a visual illustration of this process: the smaller the $\alpha$, the more likely that the encoder would predict a token label for local areas that are partially masked (marked by the red-colored grid locations in the figure). When $\alpha$ is set to be larger than 0.5, more observable pixels are ``blocked out'' and left to be predicted in the next stage. We have empirically found that setting $\alpha=0.5$ produces the best inpainting results.


\textbf{Encoder Design} Let $\image$ denotes an input image, and $\mask$ denotes a mask of the same size, given a pre-trained codebook $\codebook$ and VQGAN encoder $E_{VQ}$ of a dataset, our encoder $E(\image_{\mask})$ learns to predict the probability of token labels in each visible region of a partial image $\image_{\mask} = \image \bigodot \mask$, supervised by the ``ground-truth'' token labels at those locations from encoding the complete image with $E_{VQ}(\image)$. Our encoder is constructed by the restrictive partial convolutions and self-attention layers. It processes a partial image into probability estimations on the token labels $\Labels_{\hat{\mask}}=\{\tlabel_i\}$, given the down-sampled mask $\hat{\mask}$. The training objective is hence minimizing the negative log-likelihood:

 \begin{equation}
 \label{eq:loss_encoder}
 \mathcal{L}_{encoder} = - \mathbb{E}_{\tlabel \in \Labels} [ \sum_{\forall i \in \hat{\mask}} \text{log}\, p(y_i \vert \image_{\mask}) ],
 \end{equation}
where $\tlabel$ are the target tokens, and $\image_{\mask}$ is the partial image.

\subsection{Predicting the latent codes}
\label{sec:transformer}


The restrictive encoder has thus far encoded the input image into two distinctive regions of tokens: the visible region labeled by valid tokens $D = \{\tlabel_i\}_{\hat{\mask}}$, and the unseen region $\bar{D} = \{\mathbf{m}\}_{1-\hat{\mask}}$ that contains a set of \small [MASK] \normalsize tokens $\mathbf{m}$ (visualized by grey blocks in Figure~\ref{fig:alpha}). A bidirectional transformer based on the BERT model~\cite{devlin2018bert} is used to predict the token indices for each masked location in $\bar{D}$ based on the visible set of tokens $D$. The transformer retrieves visual features of the visible labels from the codebook, augments them with positional encoding, and processes them with attention layers to make independent label predictions on each masked location.


\textbf{Training the transformer} Training the generative transformer is simply by the maximum likelihood estimation - learning to predict the missing labels from the available ones:

 \begin{equation}
 \label{eq:loss_transformer}
 \mathcal{L}_{\text{transformer}} = - \mathbb{E}_{\tlabel \in \Labels} [ \sum_{\forall i \in \bar{D}} \text{log}\, p(y_i \vert \codebook(D)) ],
 \end{equation}

 where the only differences from Eq.\ref{eq:loss_encoder} are that the labels are predicted for the set of unseen locations $\bar{D}$ and the input to the transformer is a flattened list of visual tokens retrieved from the codebook $\codebook(D)$. We train the transformer with full images and only mask the down-sampled token map during the training. Specifically, during training, full images are encoded by the VQGAN encoder $E_{VQ}$ to obtain a complete list of tokens. The list of tokens are then randomly masked by a ratio between 15\% and 75\%.  
 

\textbf{Sampling with the transformer} During inference, the missing tokens are predicted iteratively through a parallel decoding algorithm~\cite{chang2022maskgit}. Given a sampling step $k=5$, and a cosine scheduling function $f$, the algorithm predicts labels for all missing tokens $\bar{D}$ at each step $i$, while only choosing to keep $n = f(i)$ predicted tokens with top prediction scores given by the transformer. The cosine scheduling function is chosen to ensure that $\sum_{i=0}^{k}{f(i)} = \vert \bar{D} \vert$. As pluralistic results are desired, we sample each token by drawing from its predicted probability distribution $p(y_i \vert \codebook(D))$. 

An important modification we add to the sampling algorithm from~\cite{chang2022maskgit} is the inclusion of an adaptive temperature $t$ which scales the logits prior to the softmax function with $p_i = \frac{\text{exp}(tz_i)}{\sum_n{\text{exp}(tz_i)}}$. The temperature controls the confidence level of the sampler: the lower the temperature, the more likely that labels with higher confidence scores would be sampled. Empirically, we found it beneficial to start with a high temperature and gradually anneal the temperature in each step. This encourages the sampler to introduce more diverse tokens early on and draw with more certainty when more evidence is present. In our model, we use a starting temperature of 1, and set the annealing factor $s=0.9$, which scales the temperature value in each sampling step. The chosen starting temperature has substantial impacts on the visual quality and diversity of the inpainting results (see the ablation study section in Section~\ref{sec:exp}).


\subsection{Decoding the latent codes}
\label{sec:decoder}

\begin{figure}[t]
  \centering
   \includegraphics[width=0.7\linewidth]{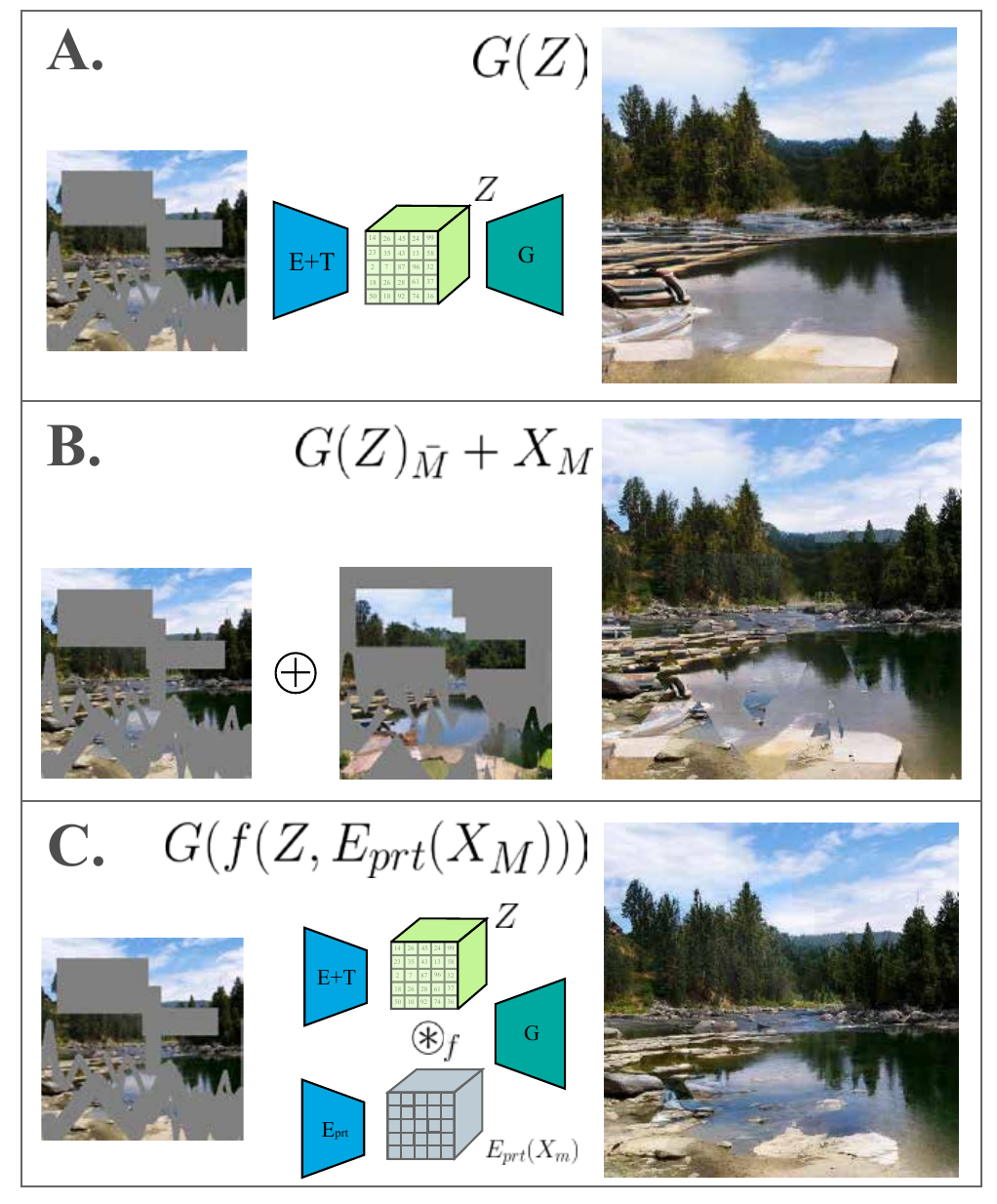}
   \caption{A visual comparison between the decoder designs. \textbf{A}. Directly decoding the predicted latent codes $Z$ with the restrictive encoder $E$ and transformer $T$, and \textbf{B}. its composition with the source image $X_M$. \textbf{C}. Our proposed decoding design, where partial image priors $E_{prt}(X_M)$ are composed with $Z$ through a composition function $f$ described in Equation.\ref{eq:decoder_1}-\ref{eq:decoder_2}.}
    \label{fig:decoder}
\end{figure}

Due to the discrete, quantized nature of the codebook representation, the visual tokens learned to encode an image usually do not fully recover the original image. Stochastically sampled tokens may further alter the global appearance of the fill-in area if they are decoded into an image directly (see Figure.~\ref{fig:decoder}.A). Therefore, compositing the generated pixels with the partial image oftentimes results in noticeable discontinuities at the mask boundaries  (see Figure~\ref{fig:decoder}.B). 

In order to synthesize pixels that coherently complete the partial images, we find it necessary to couple the quantized latent codes with smooth image priors encoded from the input partial image. Since the smooth features are responsible for locally bridging the synthesized contents and the existing contents, the encoder, denoted as $E_{prt}$, utilizes layers of the standard partial convolution (Eq.\ref{eq:partial}) to extract local features propagated to the masked regions. The features in the masked region are combined with the latent codes via an averaging operation: 
\begin{equation}
\label{eq:decoder_1}
h1 = (1-\hat{M})(Z + E_{prt}(X_M)) / 2,    
\end{equation}
where $Z$ is a feature map of the tokens predicted by the transformer. The recomposed features in the empty space are then combined with features extracted from the visible area and decoded by a convolutional generator $G$: 


\begin{equation}
\label{eq:decoder_2}
X' = G(h1 + \hat{M} E_{prt}(X_M)).
\end{equation}


The design is visualized in Figure~\ref{fig:decoder}.C. During training, the network learns to recover the ground truth image $X$ given the partial image $X_M$ and a set of chosen latent codes. As we train the network with  a reconstruction objective, we use the set of latent codes obtained from encoding the ground truth image $X$, where $Z=E_{VQ}(X)$. The encoder $E_{prt}$ and the generator $G$ are optimized by a combination of the adversarial loss~\cite{goodfellow2014generative}  $\mathcal{L}_{adv}$ with R1 regularization~\cite{ross2018improving} and the perceptual reconstruction loss (LPIPS)~\cite{zhang2018unreasonable} $\mathcal{L}_P$ with:

\begin{equation}
    \mathcal{L}_{decode} = \mathcal{L}_{adv} + 0.1 R_1 + 0.1 \mathcal{L}_P.
\end{equation}

\begin{figure*}[t]
  \centering
  \includegraphics[width=0.78\linewidth]{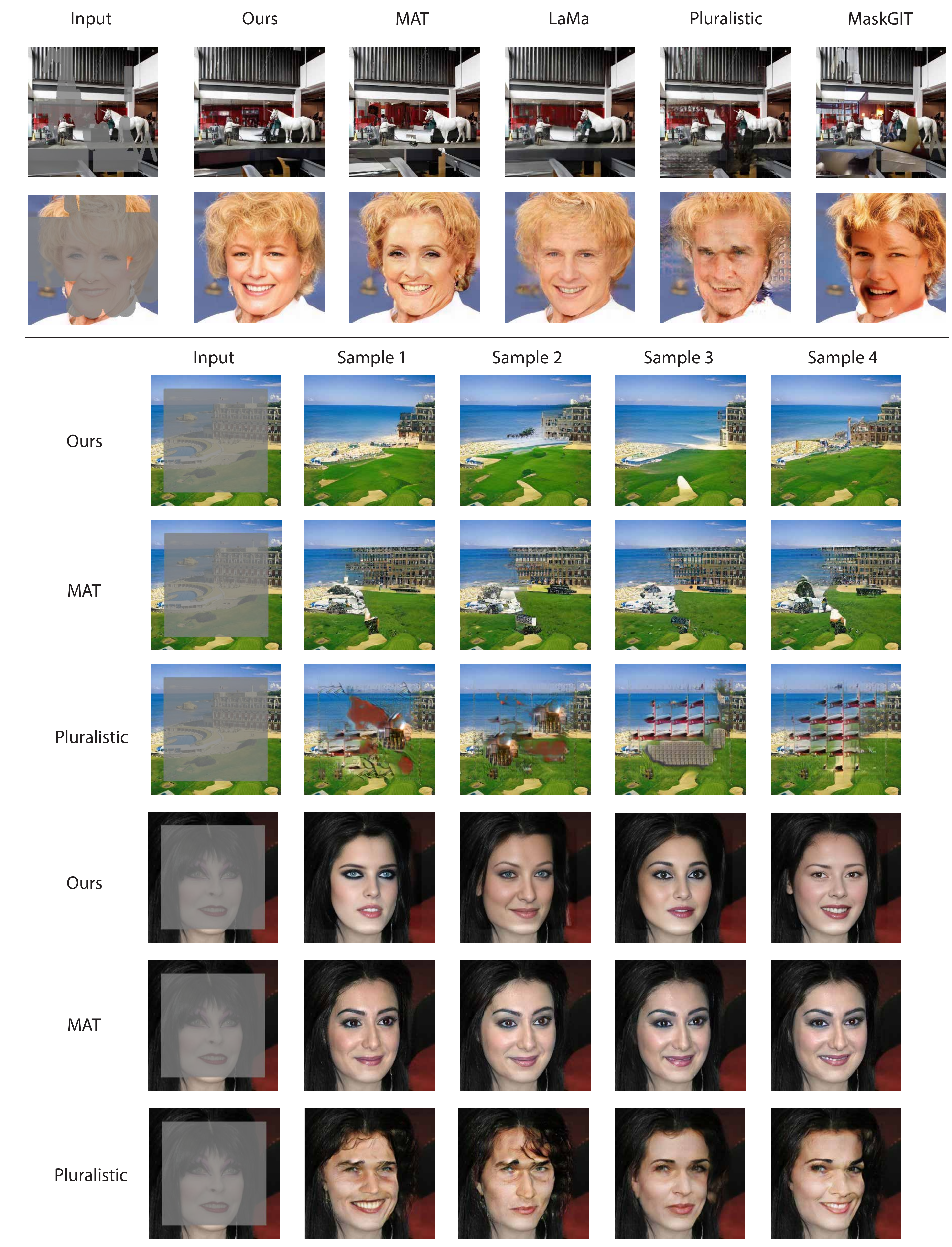}
  \caption{Visual examples on inpainting with both the random masks (upper half) and the challenging large box mask (lower half), compared to the selected baseline methods.}
  \label{fig:visualmain}
\end{figure*}

\begin{table*}[t]
\centering
\small
\begin{tabular}{l|ccc|c|ccc|c}
\hline
  \multirow{3}{*}{Methods} & \multicolumn{4}{c|}{Places ($256 \times 256$)}  & \multicolumn{4}{c}{CelebA-HQ ($256 \times 256$)} \\ \cline{2-9}
   & \multicolumn{3}{c|}{FID$\downarrow$}  & Diversity$\uparrow$ & \multicolumn{3}{c|}{FID$\downarrow$} & Diversity$\uparrow$ \\ \cline{2-9}
       & Small Mask & Large Mask &  Box  & Box & Small Mask & Large Mask & Box & Box  \\ \hline 
Ours & \textbf{1.02} & \textbf{2.82}    & \textbf{13.30} & \textcolor{blue}{0.29$\pm$0.06}   & \textbf{2.70}        & \textbf{5.04}    & \textbf{12.79} & \textbf{0.28$\pm$0.05}  \\ \hline
MAT\cite{li2022mat}  & \textcolor{blue}{1.19} & \textcolor{blue}{3.32}    & \textcolor{blue}{17.5} & 0.26 $\pm$ 0.04   & \textcolor{blue}{2.94}        & \textcolor{blue}{5.16}           & \textcolor{blue}{15.18} & 0.10 $\pm$ 0.02  \\  
LaMa\cite{suvorov2021resolution}  & 1.22 & 3.78    & 19.48 & -   & 3.98        & 8.75           & 23.24 & - \\ 
MaskGIT\cite{chang2022maskgit} & 19.84 & 36.38    & 52.71 & \textbf{0.39$\pm$0.05}   & 19.68        & 40.76            & 30.87 &  \textcolor{blue}{0.25$\pm$0.04} \\  
Pluralistic\cite{zheng2019pluralistic} & 4.83 & 16.26    & 86.57 & -   & 9.7        & 28.89            & 43.08 & 0.18$\pm$0.02 \\   \hline
\end{tabular}
\caption{Comparisons of FID and diversity scores to the baseline methods. \textbf{Bold} text denotes the best, and \textcolor{blue}{blue} text denotes the second. Since LaMa~\cite{suvorov2021resolution} does not generate pluralistic results, and Pluralistic~\cite{zheng2019pluralistic} produces degenerate results in the Places Box setting, we omit their diversity scores in the table.
}
\label{tab:main}
\end{table*}

\section{Experiments}
\label{sec:exp}
\textbf{Dataset} Our experiments are conducted on the Places365-Standard~\cite{zhou2017places} and the CelebA-HQ~\cite{karras2018progressive}, two benchmarks widely evaluated by past image inpainting methods. The Places365-Standard dataset contains 1.8 million images for training and 36.5 thousand images for evaluation across over 205 scene categories. The CelebA-HQ dataset is split into 24,183 training images and 2,993 test images. For both datasets, we use an image resolution of $256\times256$. We further set three different mask settings for our experiments: 1) small random hole, 2) large random hole, and 3) large box hole. The first two settings are directly adopted from MAT~\cite{li2022mat} (free-form holes with strokes and boxes). The third, challenging setting uses a very large box mask centered in the image with its width and length equal to 80\% of the image size. As such setting leaves a majority of the pixels empty, we find it suitable for the evaluation of pluralistic inpainting, as well as on whether the inpainting method extends to the extreme cases.


\textbf{Evaluation metric} Our main objectives are to evaluate both the visual quality and the sample diversity of the inpainted image. To this end, we opt for the perceptual metric FID~\cite{heusel2017gans} and the LPIPS-based diversity score~\cite{zhu2017toward}. Specifically, to compute the diversity score for each dataset, we use a smaller subset of both Places-Standard and CelebA-HQ with 1000 images. For each image, 100 inpainting samples are drawn under the large box setting, and the diversity score for each sample is computed as the average of the pair-wise LPIPS distances between the drawn samples. The final score shows the average of the individual scores and their standard deviation. In addition, we provide visual examples for a qualitative evaluation (see Figure~\ref{fig:visualmain},\ref{fig:ablationt} and more in the Supplements). 

\subsection{Comparisons to The State of Arts} 

Our results are compared to four recent baseline image inpainting methods. The baseline methods are evaluated directly with the provided pre-trained models and their public source codes: 1) \textbf{MAT}~\cite{li2022mat} is the current state-of-the-art method in image inpainting that is able to produce high-quality and pluralistic inpainting results in challenging settings. 2) \textbf{LaMa}~\cite{suvorov2021resolution} is another state-of-the-art method that is characterized by the use of Fourier convolution. However, it does not generate pluralistic results; 3) \textbf{MaskGIT}~\cite{chang2022maskgit} is a latent-code-based image synthesis method that has strongly motivated our work and has been shown to be adaptable to image inpainting; and 4) \textbf{Pluralistic}~\cite{zheng2019pluralistic} is a seminal work in exploring pluralistic image inpainting by coupling smooth latent priors with the partial image features.


Table~\ref{tab:main} lists the quantitative evaluation results: ours outperforms the baseline methods in all settings in terms of FID and diversity score, except for the diversity score on Places, where we rank the second best. While MaskGIT produces more diverse results on the Places dataset, it does so at the cost of visual quality, as the method is not designed to coherently compose the synthesized contents with the existing contents. Figure~\ref{fig:visualmain} shows visual examples of the inpainting results under the three different mask settings. While our results can be seen slightly better in inpainting consistency compared to the baselines, the method truly shines in the pluralistic comparisons (lower half of the figure): while our model synthesizes pluralistic inpainting results that are high-quality and varied in both local details and global structures, the baseline methods either only produce locally diversifies results (MAT~\cite{li2022mat}, see 4th,7th rows in Figure~\ref{fig:visualmain}) or fail to generate visually coherent results in the challenging box setting (Pluralistic~\cite{zheng2019pluralistic}, see 5th,8th rows in Figure~\ref{fig:visualmain}).

\begin{table}[t]
\centering
\small
\begin{tabular}{l|c|c}
\hline
Type & Model & FID$\downarrow$ \\ \hline
 & Full Model & \textbf{12.81} \\ \hline
\multirow{3}{*}{Temperature} & $t=0.1$ & 14.62 \\ 
 & $t=0.5$ & 13.61 \\
 & $t=2$ & 13.96 \\ \hline
\multirow{3}{*}{Restrictive Conv} & $\alpha=0.25$ & 12.9 \\ 
 & $\alpha=0.75$ & 14.12 \\ 
 & $\alpha=1.0$ & 15.04 \\ \hline
\multirow{2}{*}{Network Design}  & Vanilla Encoder & 15.32 \\ 
 & Vanilla Decoder & 18.50 \\ \hline
\end{tabular}
\caption{Quantitative ablation study. ``Temperature'' adjustments change the temperature value in the sampling procedure. ``Restrictive Conv'' adjustments change the mask update rule in the restrictive encoder. ``Network Design'' adjustments replace our designed network structures with the vanilla ones: for the ``Vanilla Encoder'' setting, an encoder network with the regular convolution layers are used; for the ``Vanilla Decoder'' setting, the predicted latent codes are directly decoded into an image. In the ``Full Model'', we set $t = 1.0$ and $\alpha = 0.5$.
}
\label{tab:ablation}
\end{table}

\subsection{Ablation Study} 

We validate several design choices of our model with an ablation study on a smaller evaluation subset of the Places dataset with 3000 images. Table~\ref{tab:ablation} shows the quantitative results of the ablation study based on the the FID score metric.

\textbf{Effectiveness of the Restrictive Design} One main question we ask when designing the restrictive encoder is how much our model needs to separate small mask regions. In the extreme case, a one-pixel mask can be turned into a masked token in the down-sampled token grid, thus discarding 93.74\% of information around that pixel. Obviously, this behavior is not desirable and we have experimentally found that larger $\alpha$ values lead to decreased performance (see Table~\ref{tab:ablation} and Figure~\ref{fig:alpha}). Our final choice of $\alpha=0.5$ leads our encoder to complete the inpainting by itself for local regions that are less than 50\% masked. On average, the restrictive encoder achieves a label classification accuracy of 23\% for labels in small mask regions and near mask boundaries in the large mask setting (tested on the Places dataset), whereas an encoder with the regular convolution only achieves an accuracy of 9\%. Furthermore, due to the inductive bias learned in the training process, a ``wrongly'' classified label does not necessarily translate to poor inpainting results. The result could be a completed image that is different from the original one, while still visually plausible. We provide some further visual comparisons with regard to this observation in the supplementary material.

\begin{figure}[t]
  \centering
  \includegraphics[width=1.0\linewidth]{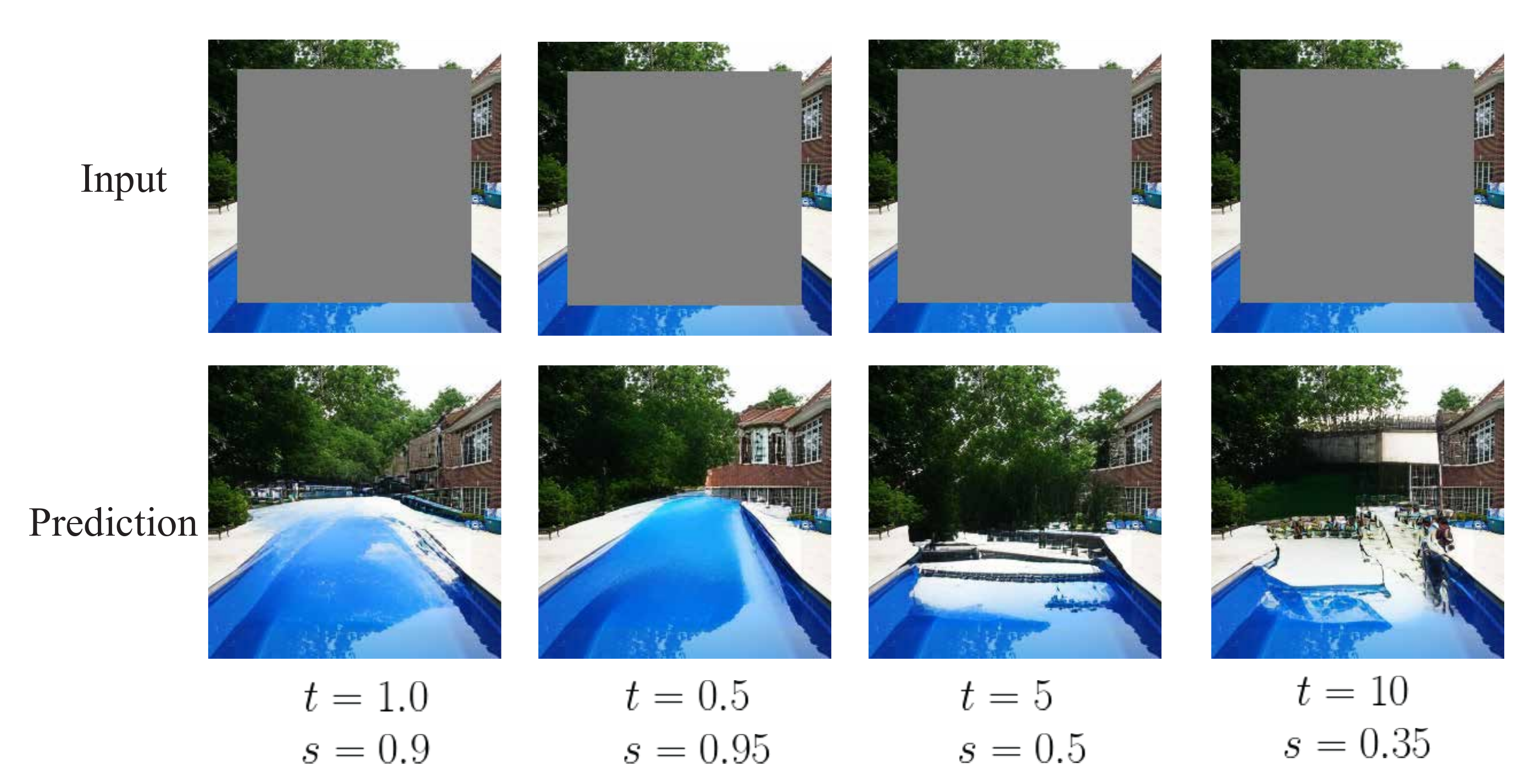}
  \caption{Comparisons of inpainting results with regard to different sampling temperature $t$ and annealing factors $s$.}
  \label{fig:ablationt}
\end{figure}

\textbf{Effectiveness of the Sampling Function} As described in Section~\ref{sec:transformer}, we have found the adaptive temperature a key factor in controlling the visual quality and diversity outcome of the inpainting. Figure~\ref{fig:ablationt} provides a visual comparison between different configurations of the temperature and annealing factor. We have observed that a larger starting temperature creates more diverse output, though at the risk of breaking the coherence of the inpainting. Overly small temperature, on the other hand, led our model to mainly interpolate patterns in the masked area, resulting in large areas of homogeneous textures in the inpainting results. More visual comparisons are included in the supplementary material.

\begin{figure}[t]
  \centering
  \includegraphics[width=0.85\linewidth]{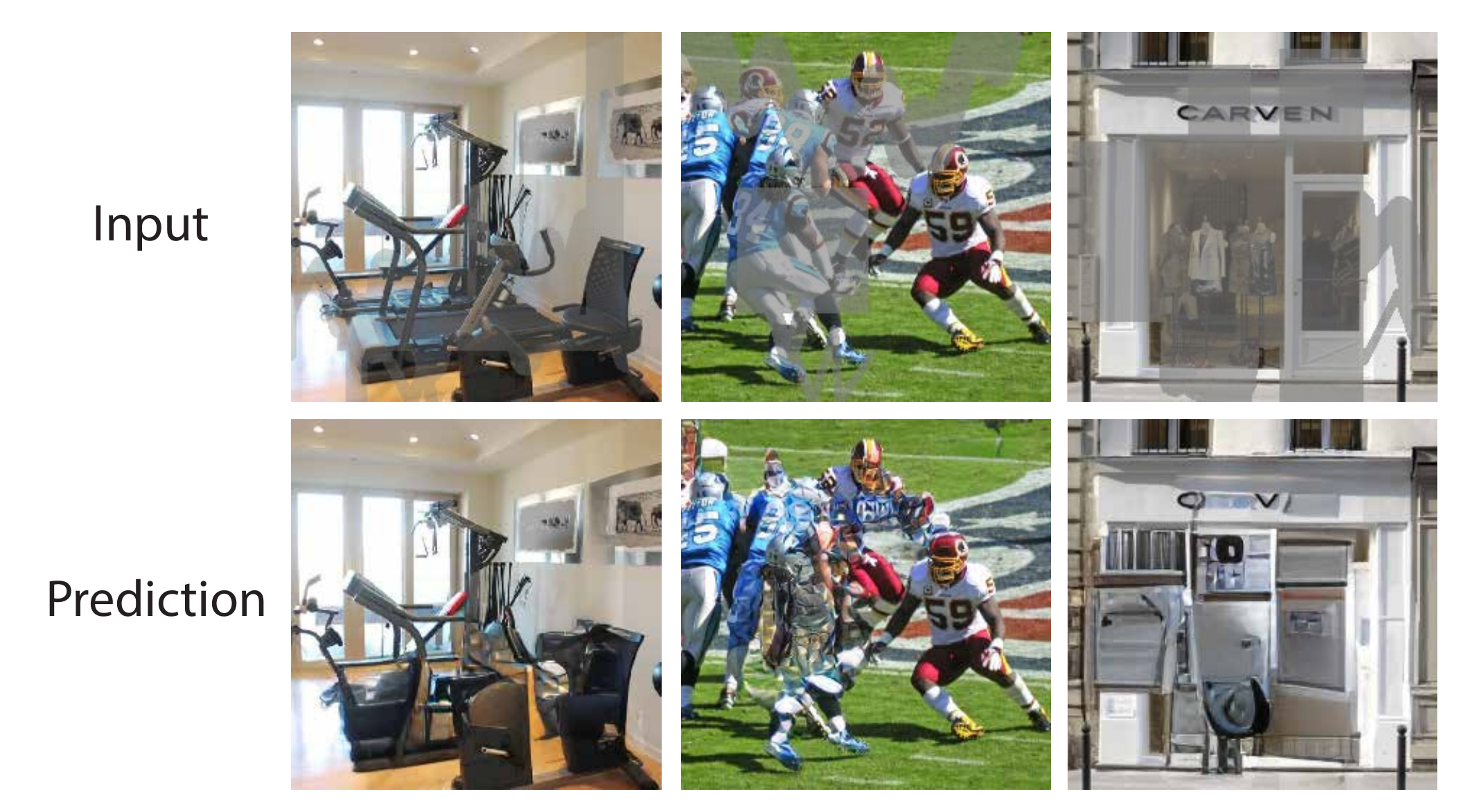}
  \caption{Failure cases in our results.}
  \label{fig:failure}
\end{figure}

\subsection{Limitations of our model} 

Our model shares limitations with previous inpainting methods in its limited ability to complete semantically salient objects such as shops and furniture, as well as people and animals, when trained on the Places dataset (see Figure~\ref{fig:failure}). The inference speed of our method is also relatively slower than the end-to-end, single-pass inpainting methods, as the majority of the computation time is spent on the iterative sampling steps (see supplementary material for an inference time comparison). We also have left some areas unexplored. For instance: whether the aforementioned synthesis problem can be mitigated by training with semantic labels, or how the model extends to higher resolution input. We believe that these extensions are well within reach as past latent code synthesis methods (e.g.~\cite{esser2021taming, chang2022maskgit}) have demonstrated such capabilities.


\section{Conclusion}

In this paper, we present a pluralistic image inpainting method that first analyzes only the visible and near-visible regions through latent code prediction, and synthesizes the missing contents through a versatile bidirectional transformer and a reconstruction network that composes the code prediction with partial image priors. We have validated our design choices through comparative experiments on public benchmarks and an ablation study, as our method achieves state-of-the-art performance in both visual quality and sample diversity.


\section*{Acknowledgements}
This research was sponsored by the Army Research Office and was accomplished under Cooperative Agreement Number W911NF-20-2-0053, and by the U.S. Army Research Laboratory (ARL) under contract number W911NF-14-D-0005. This work was also supported by the Office of Naval Research under grant N00014-22-1-2020. The views and conclusions contained in this document are those of the authors and should not be interpreted as representing the official policies, either expressed or implied, of the Army Research Office or the U.S. Government. The U.S. Government is authorized to reproduce and distribute reprints for Government purposes notwithstanding any copyright notation.
\clearpage

{
    \small
    \bibliographystyle{ieeenat_fullname}
    \bibliography{main}
}

\clearpage
\setcounter{page}{1}

\twocolumn[{%
\renewcommand\twocolumn[1][]{#1}%
\maketitlesupplementary
\begin{center}
    \centering
    \captionsetup{type=figure}
    \includegraphics[width=0.8\textwidth]{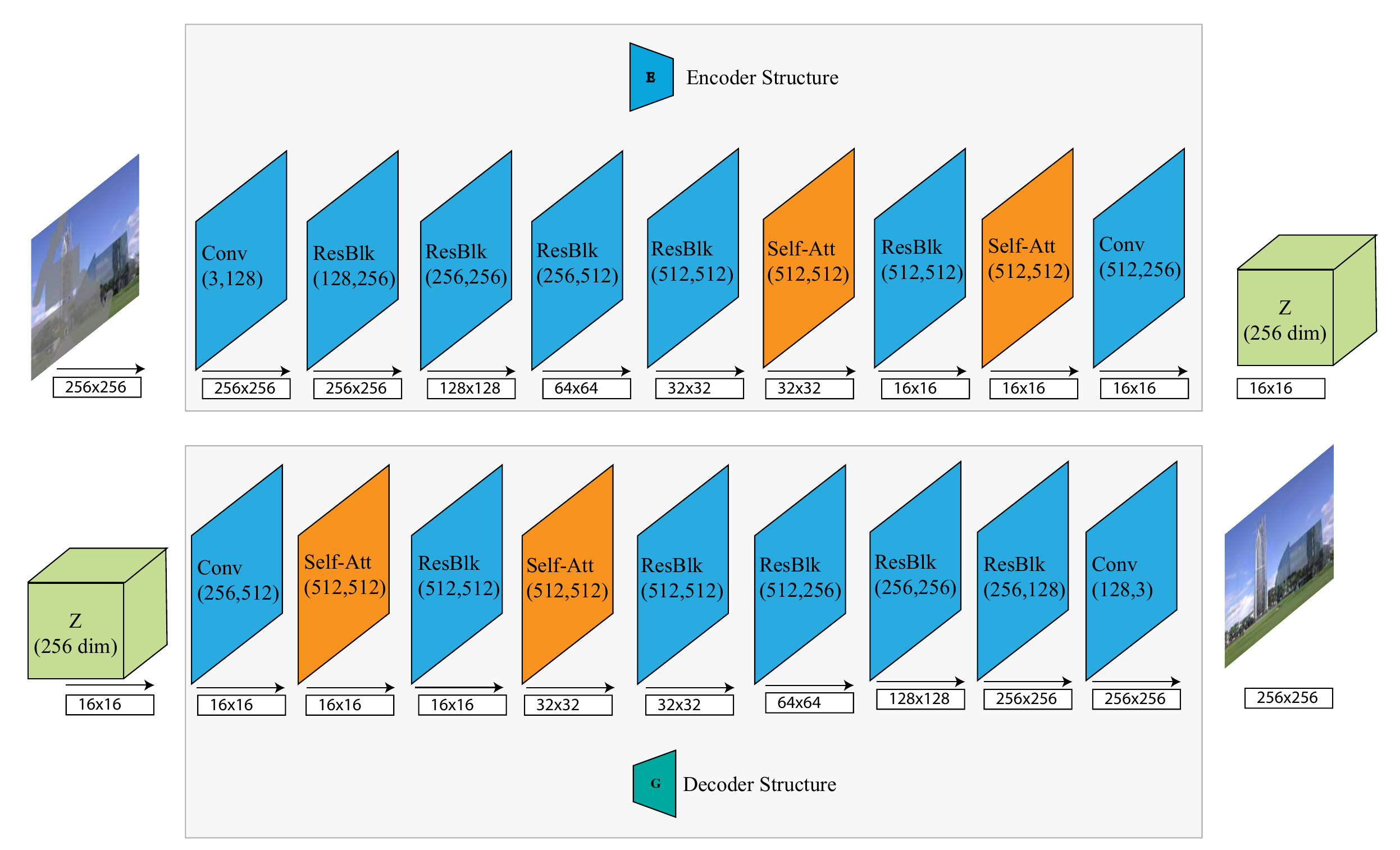}
    \captionof{figure}{Detailed network structures for the encoder and decoder. Numbers within each feature map (e.g. (3,128)) denote the input and output channels. Numbers below each feature map (e.g. 256x256) denotes the size of the tensor. }
    \label{fig:suppnetwork}
\end{center}}]

\section{Implementation Details} 

The network structures and all hyper-parameter settings are identical for both the Places model and the CelebA-HQ model. Our training pipeline is divided into three separate stages: the training of the encoder, the transformer, and the decoder. The free-form random masks~\cite{li2022mat} are used during the training of the encoder and the decoder. For the pre-trained VQGAN model, we use a quantization of $N=1024$ tokens, each with a 256-channel embedding. All models are trained on 3 NVidia V100 GPUs with a batch size of 8. The Places model is trained for 20 epochs, and the CelebA-HQ model is trained for 200 epochs. The inference time of our model is around 0.4 seconds for an image on a single NVidia V100 GPU regardless of the mask size.

Figure~\ref{fig:suppnetwork} shows the detailed network structure designs for the encoders and decoder used in our method. Given input channel $c_{in}$ and output channel $c_{out}$, ``ResBlk'' is a resnet block composed of two convolution layers and a skip connection. The two convolutions have weight matrices $W_1 \in \Real^{3 \times 3 \times c_{in} \times c_{out}}$ and  $W_2 \in \Real^{3 \times 3 \times c_{out} \times c_{out}}$ respectively. Layer normalization is applied before and after the first convolution. The restrictive encoder and partial encoder described in the method part (Section~\ref{sec:method}) replace all convolutions with the restrictive convolution and the partical convolution respectively. In addition, self-attention layers (denoted by ``Self-Att'' in the figure) are added to process the features at 32x32 and 16x16 resolution. 

The transformer model described in Section~\ref{sec:transformer} is designed based on the minGPT transformer model\footnote{see \href{https://github.com/karpathy/minGPT}{https://github.com/karpathy/minGPT}}, with token embedding and positional embedding of 1408 channel and 40 layers of 16-head attention layers. During training, we have applied attention dropout and embedding dropout both with a 10\% probabilities.






\section{Detailed Loss Function}

The loss functions used in training the decoder network described in Section~\ref{sec:decoder} involve an adversarial loss function~\cite{goodfellow2014generative}:

\begin{align}
    \mathcal{L}_G &= - \mathbb{E}_{\hat{x}}[\text{log}(D(\hat{x})], \\
    \mathcal{L}_D &= - \mathbb{E}_{x}[\text{log}(D(x)] - \mathbb{E}_{\hat{x}}[\text{log}(1 - D(\hat{x})], 
\end{align}

where $x$ and $\hat{x}$ are a pair of real and fake samples, $G$, $D$ are the generator and the discriminator. We additional use the R1 regularization~\cite{ross2018improving} of the form:

\begin{equation}
    R_1 = \mathbb{E}_{x} \| \triangledown D(x) \|.
\end{equation}

The LPIPS reconstruction loss function~\cite{zhang2018unreasonable} is formulated as

\begin{equation}
    \mathcal{L}_P = \sum_{l} \frac{1}{H_l W_l} \sum_{h,w} \| w_l \odot (\hat{y}^{l}_{hw} - \hat{y}^{l}_{0hw} \|_{2}^{2},
\end{equation}

which compute the L2 distance between the layer activation of a pretrained VGG network~\cite{simonyan2014very} at each layer $l$.

\begin{figure*}[t]
  \centering
  \includegraphics[width=0.95\linewidth]{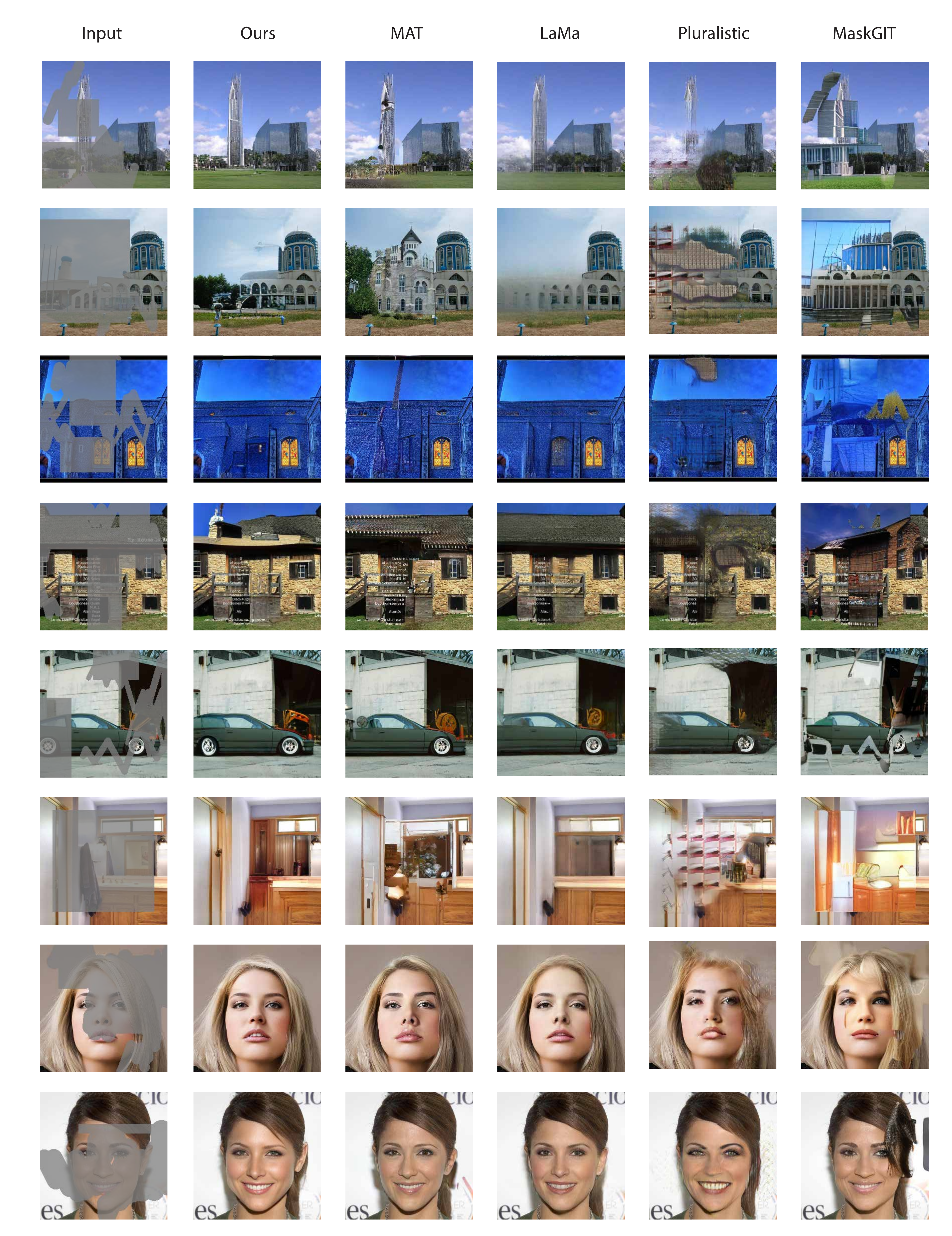}
  \caption{Further visual examples of inpaining under the large mask setting, compared to the baseline methods.}
  \label{fig:suppmain1}
\end{figure*}

\begin{figure*}[t]
  \centering
  \includegraphics[width=0.8\linewidth]{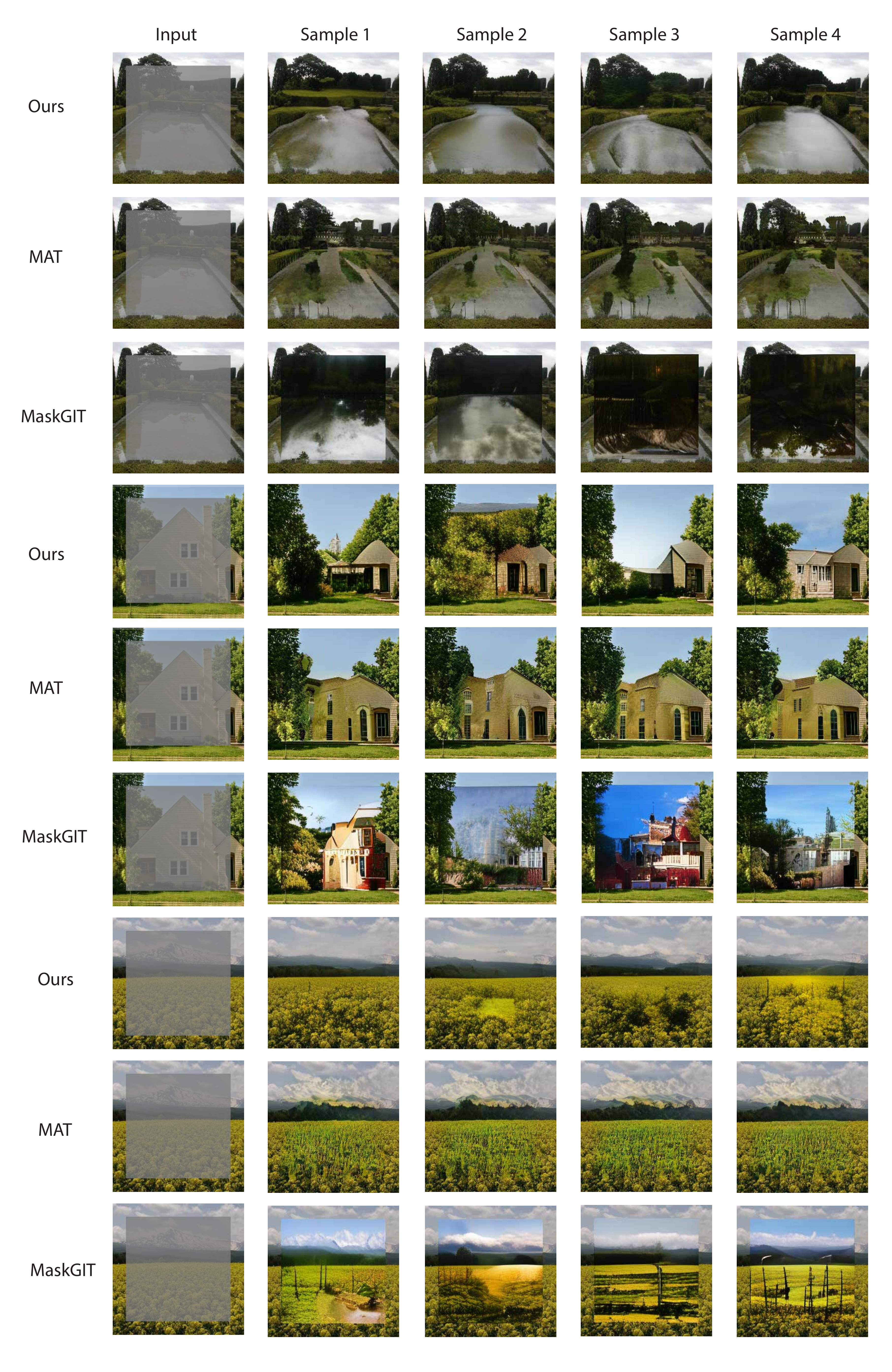}
  \caption{Further visual examples of pluralistic inpainting on the Places Dataset~\cite{zhou2017places}, compared to the baseline methods.}
  \label{fig:suppmain2}
\end{figure*}

\begin{figure*}[t]
  \centering
  \includegraphics[width=0.8\linewidth]{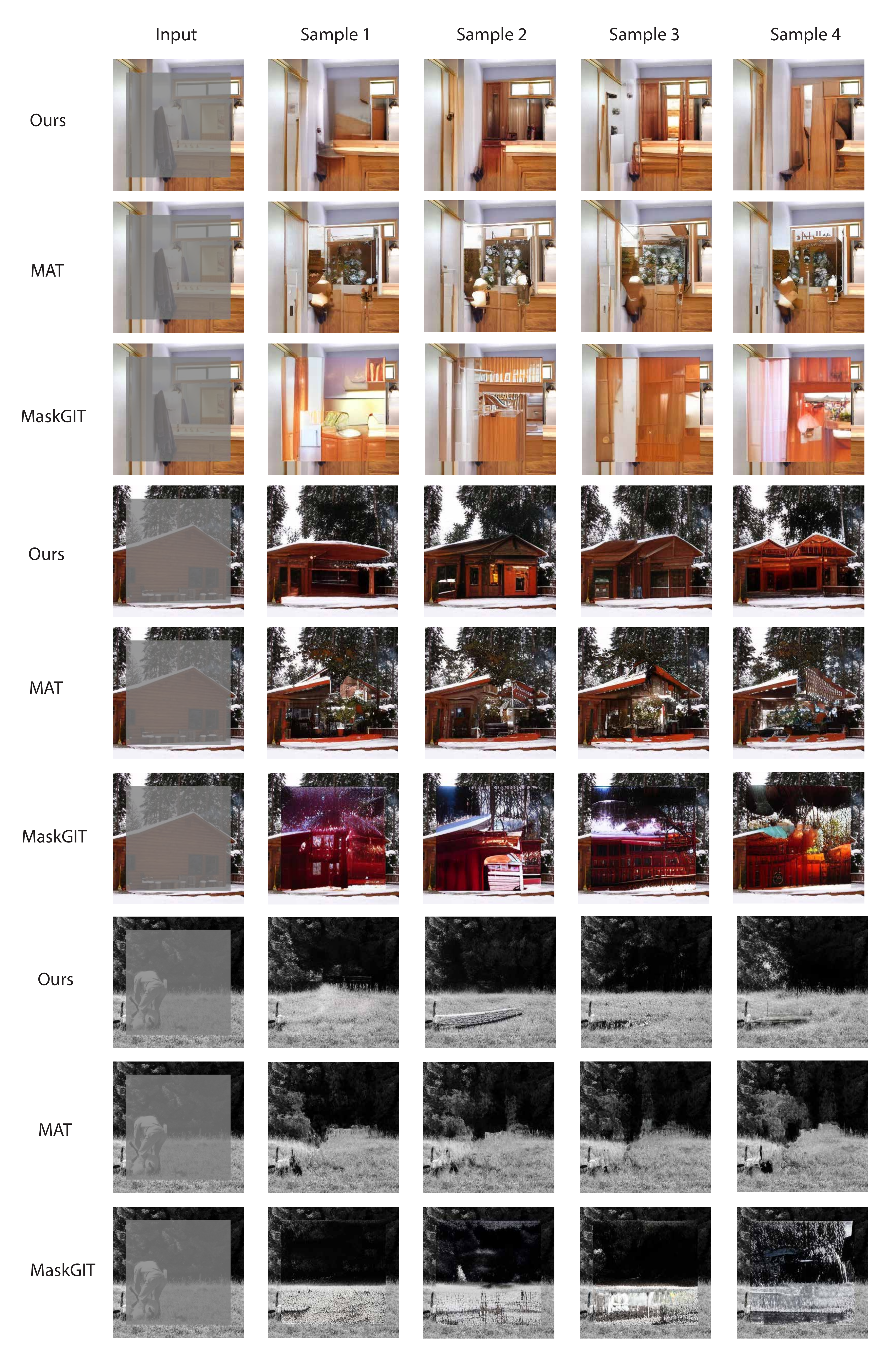}
  \caption{Further visual examples of pluralistic inpainting on the Places Dataset~\cite{zhou2017places}, compared to the baseline methods.}
  \label{fig:suppmain3}
\end{figure*}

\begin{figure*}[t]
  \centering
  \includegraphics[width=0.8\linewidth]{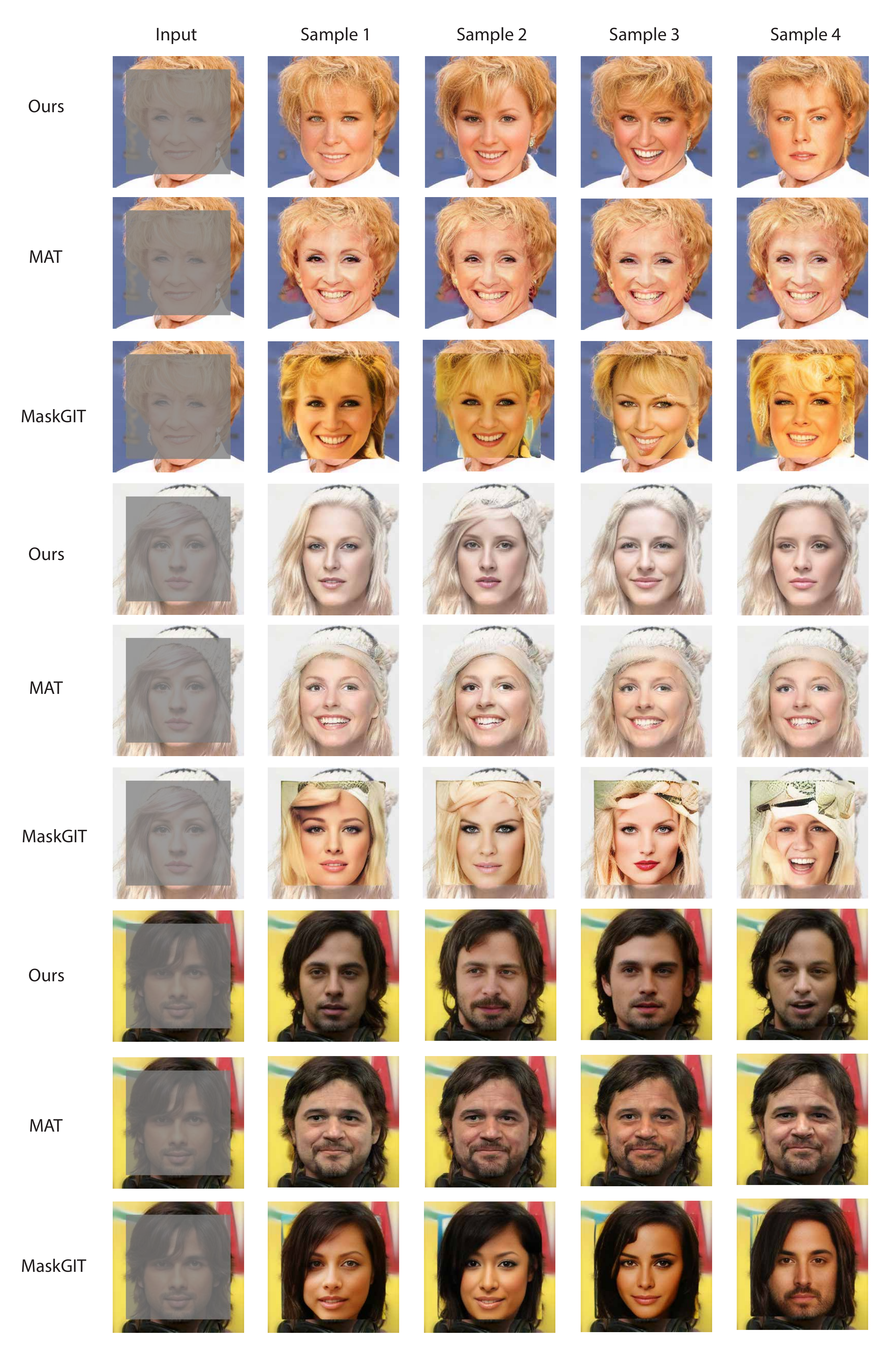}
  \caption{Further visual examples of pluralistic inpainting on the CelebA-HQ Dataset~\cite{karras2018progressive}, compared to the baseline methods.}
  \label{fig:suppmain4}
\end{figure*}

\begin{figure*}[t]
  \centering
  \includegraphics[width=0.7\linewidth]{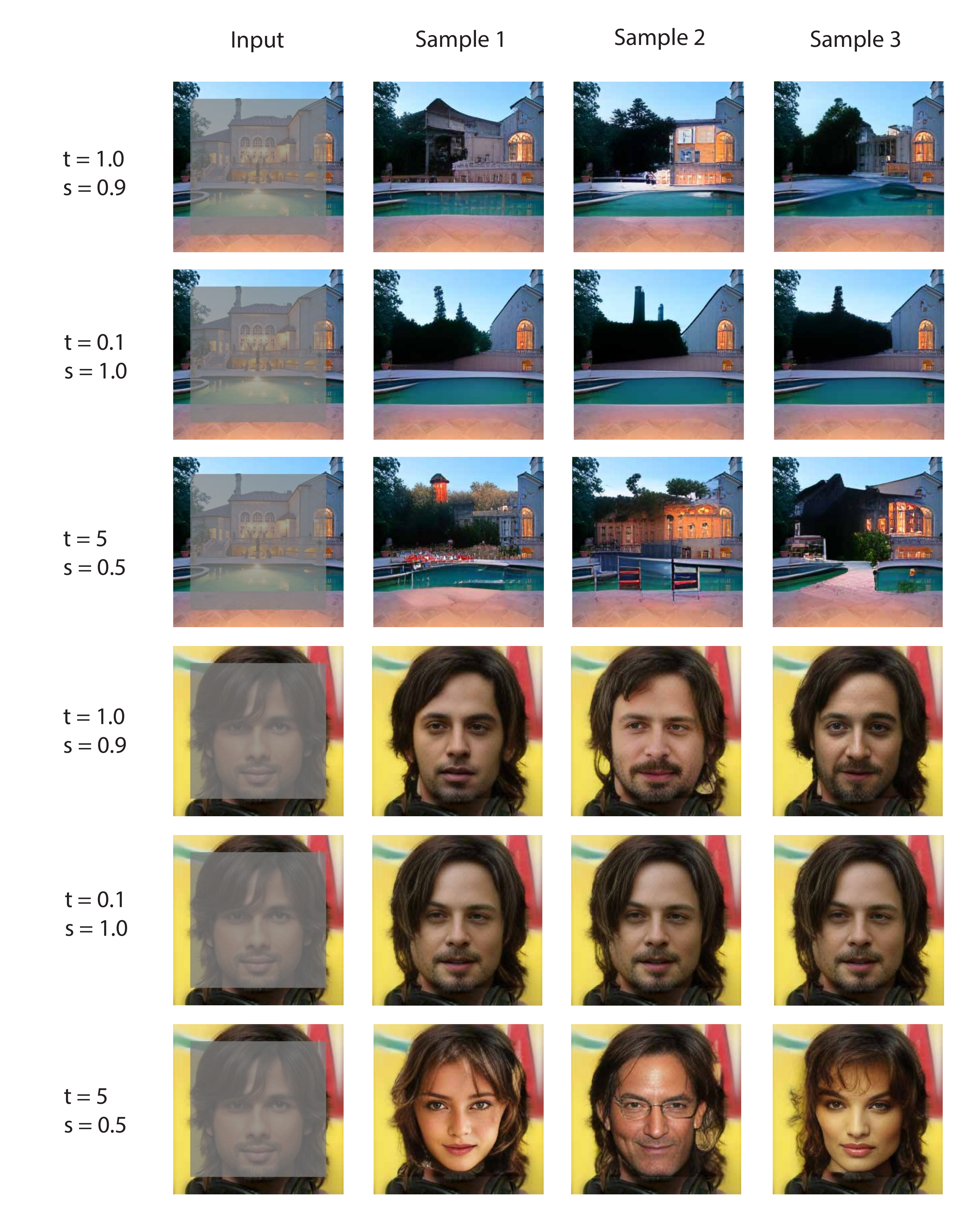}
  \caption{Further visual examples of pluralistic inpainting with respect to different sampling temperature $t$ and annealing factor $s$.}
  \label{fig:supptemp}
\end{figure*}

\section{Additional Visual Results}
Please refer to Figure~\ref{fig:suppmain1}, \ref{fig:suppmain2}, \ref{fig:suppmain3}, \ref{fig:suppmain4} for additional visual comparisons to the baseline methods on both the Places~\cite{zhou2017places} and the CelebA-HQ~\cite{karras2018progressive} dataset.

\begin{figure*}[!tbhp]
  \centering
  \includegraphics[width=0.8\linewidth]{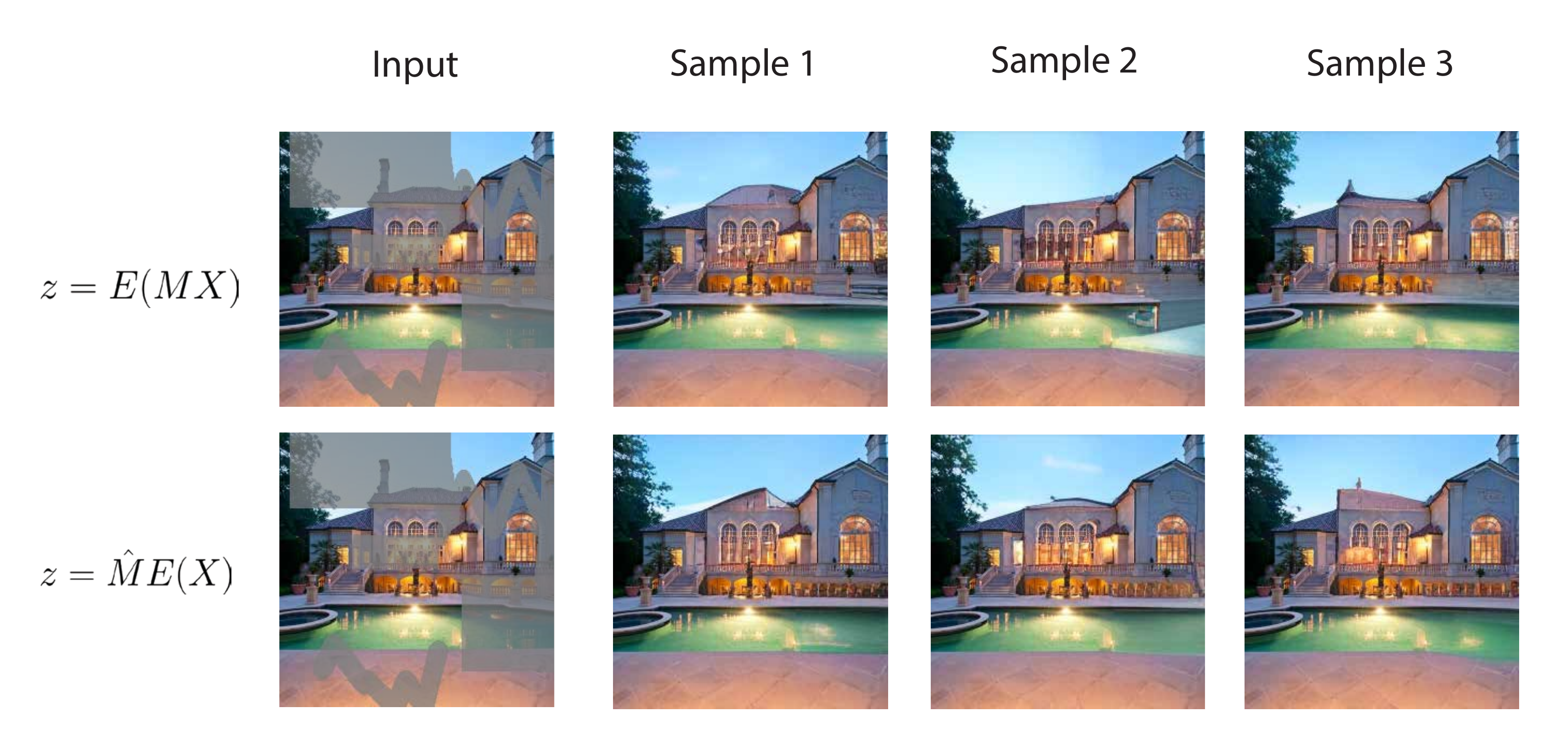}
  \caption{Visual comparison between inpainting with the restrictive encoder and a miracle encoder.}
  \label{fig:suppencoder}
\end{figure*}

\section{Further ablation analysis}
\textbf{Ablation study on the sampling function.} Figure~\ref{fig:supptemp} provides more visual examples that compare between inpainting results under different sampling temperature $t$ and annealing factor $s$ (described in Section~\ref{sec:transformer}). As suggested by the examples, lower temperature in the sampling function leads to less diverse results and homogeneous textures in the synthesized areas (particular in natural scene inpainting). Higher temperature, on the other hand, leads to more diverse results at the expanse of visual coherence. The bottom three rows of Figure~\ref{fig:supptemp} best illustrates this phenomenon: the small mustache region left in the original masked image should naturally encourage the inpainting algorithm to synthesize a male face. However, since setting a higher temperature encourages the sampler to sample random latent codes in the early steps, female faces are synthesized regardless of the present evidence. 

\textbf{Ablation study on the restrictive encoder.} We further study the effectiveness of the restrictive encoder by comparing it to a miracle encoder that has access to the original complete image. Specifically, while our restrictive encoder takes as input the masked image and produces latent code by $z_0 = E(MX)$, the miracle encoder encodes the complete image as input and masks the latent codes with a down-sampled mask, with $z_1 = \hat{M}E(X)$. Figure~\ref{fig:suppencoder} provides a visual comparison between the two encoders. For the specific example in the figure, $z_0$ and $z_1$ only share 24.8\% of the encoded latent codes, as the miracle encoder manages to encode the image differently with access to the complete image. The quality of the inpainting results, however, show little difference, although the restrictive encoder has to infer latent codes with far less information. Quantitative evaluation in Table~\ref{tab:supp} provides a comparison between the performance of the restrictive encoder and the miracle encoder when used in the full inpainting pipeline, where we found that the designed restrictive encoder can be nearly as effective as one that has full access to the complete images. This indicates that meaningful inductive bias has been learnt by the encoder in the training process.

\begin{table}[t]
\centering
\small
\begin{tabular}{|l|cc|c|}
\hline
  \multirow{3}{*}{Methods} & \multicolumn{3}{c|}{Places ($256 \times 256$)}  \\ \cline{2-4}
   & \multicolumn{2}{c|}{FID$\downarrow$}  & Diversity$\uparrow$  \\ \cline{2-4}
       & Small Mask & Large Mask &  Box  \\ \hline 
Restrictive & 1.02 & 2.82   & 0.29$\pm$0.06    \\ \hline
Miracle  & 0.93 & 2.71    & 0.29$\pm$0.06    \\  
  \hline
\end{tabular}
\caption{Comparisons of FID and diversity scores between the restrictive encoder and the miracle encoder.}
\label{tab:supp}
\end{table}

\section{Time and Memory Complexity} Our model takes 314ms to generate an $256 \times 256$ inpainted image, with 25ms on encoding, 61ms on decoding and 228ms on predicting tokens, which is similar to the speed of the generative-transformer-based method MaskGIT [4] (298ms), but slower than the one-pass inpainting methods such as LaMa [34] (86ms) and MAT [22] (83ms). Inference with our model takes 5893Mb of GPU memory to process a single image.

\end{document}